\documentclass[10pt,twocolumn,letterpaper]{article}

\usepackage[pagenumbers]{wacv} 

\usepackage{graphicx}
\usepackage{amsmath}
\usepackage{amssymb}
\usepackage{booktabs}

\usepackage{multirow}
\usepackage{bm}
\usepackage{algorithmic}
\usepackage{algorithm}
\usepackage[algo2e,linesnumbered]{algorithm2e}
\usepackage{color}
\usepackage[table]{xcolor}
\newcommand{\red}[1]{\textcolor{red}{#1}}
\newcommand{\blue}[1]{\textcolor{blue}{#1}}  
\newcommand{\round}[1]{\ensuremath{\lfloor#1\rceil}}

\usepackage[pagebackref,breaklinks,colorlinks]{hyperref}

\usepackage[capitalize]{cleveref}
\crefname{section}{Sec.}{Secs.}
\Crefname{section}{Section}{Sections}
\Crefname{table}{Table}{Tables}
\crefname{table}{Tab.}{Tabs.}

\begin{document}

\title{Self-Relaxed Joint Training: \\ Sample Selection for Severity Estimation with Ordinal Noisy Labels}

\author{
Shumpei Takezaki$^{1}$
\and
Kiyohito Tanaka$^{2}$
\and
Seiichi Uchida$^{1}$
\and
{$^{1}$Kyushu University, Fukuoka, Japan}
\space
{$^{2}$Kyoto Second Red Cross Hospital, Kyoto, Japan}
\\
{\tt\small shumpei.takezaki@human.ait.kyushu-u.ac.jp}
}

\maketitle

\begin{abstract}

Severity level estimation is a crucial task in medical image diagnosis. However, accurately assigning severity class labels to individual images is very costly and challenging. Consequently, the attached labels tend to be noisy. In this paper, we propose a new framework for training with ``ordinal'' noisy labels. Since severity levels have an ordinal relationship, we can leverage this to train a classifier while mitigating the negative effects of noisy labels. Our framework uses two techniques: clean sample selection and dual-network architecture. A technical highlight of our approach is the use of soft labels derived from noisy hard labels. By appropriately using the soft and hard labels in the two techniques, we achieve more accurate sample selection and robust network training. The proposed method outperforms various state-of-the-art methods in experiments using two endoscopic ulcerative colitis (UC) datasets and a retinal Diabetic Retinopathy (DR) dataset. Our codes are available at \url{https://github.com/shumpei-takezaki/Self-Relaxed-Joint-Training}.
\end{abstract}


\section{Introduction}\label{sec:intro}

Estimating severity levels is one of the important tasks in medical image diagnosis.
Traditionally, medical experts evaluate severity levels as discrete {\em ordinal} labels. 
For example, the severity levels of endoscopic ulcerative colitis (UC) images are evaluated as four-level Mayo scores~\cite{uc}; Mayo 0 is a normal or inactive disease, and Mayo 3 is a severe disease.\par

A practical and serious problem in estimating severity levels by machine-learning models is that ordinal labels attached to individual images tend to be noisy. This is because the severity is continuous, and many ambiguous cases exist. For example, for an image with a severity level of around 1.5, one expert might label it 1 and another 2. As a possible remedy, several public datasets, such as LIMUC~\cite{limuc} for UC, employ multiple medical experts to attach reliable labels. This remedy, however, requires enormous efforts from multiple experts. Furthermore, it is impractical for diagnostic applications in rare diseases where only a few experts exist.\par

\begin{figure*}[t]
    \centering
    \includegraphics[width=0.7\linewidth]{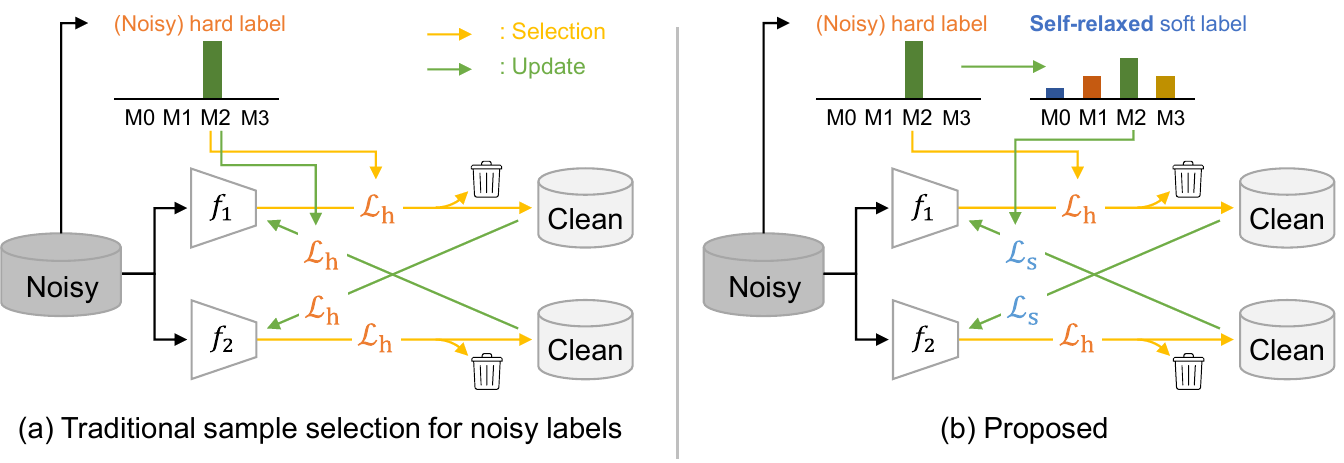}\vspace{-5pt}
    \caption{ (a) Traditional joint-training framework with dual-network model with sample selection for noisy labels. (b) The proposed {\em self-relaxed joint training} framework for learning with ordinal 
    noisy labels. ``M0’' stands for ``Mayo 0.''}
    \label{fig:overview}
    \vspace{-10pt}
\end{figure*}

A more efficient remedy to the noisy-label problem is to utilize some learning models with noisy labels~\cite{noisy_label_survey_2,Liang_2022,noisy_label_survey}. 
Fig.~\ref{fig:overview}(a) shows a traditional joint-training framework for noisy labels. This framework employs two techniques and has experimentally demonstrated its high performance.
First, the joint-training framework employs sample selection for discarding images with suspicious labels. Specifically, an image with a class label $c$ is discarded when a loss by $c$ shows a high value. Second, the framework employs a dual-network architecture. Roughly speaking, two networks $f_1$ and $f_2$ are trained jointly in a complementary manner by referring to each other's loss values. This joint-training framework is useful in canceling the negative effect of noisy labels. Note that in the traditional framework, both techniques use the same criterion $\mathcal{L}_\mathrm{h}$, which is a loss value evaluated by (noisy) hard labels (i.e., one-hot teacher vectors).\par

The traditional joint-training framework of Fig.~\ref{fig:overview}(a) still has two issues to be solved for ordinal noisy labels. The first issue is that sample selection by the loss $\mathcal{L}_\mathrm{h}$ might be imperfect. In fact, the past attempts, such as Co-teaching~\cite{co_teaching}, JoCor~\cite{jocor}, and CoDis~\cite{codis}, experimentally proved that their label precision (i.e., the ratio of samples with correct labels in the selected samples) is far below 100\%. If we cannot discard samples with incorrect labels, they harm updating the network models because they force the model to output the incorrect class.
\par

The second issue is that the traditional framework assumes non-specific noisy labels; in other words, it does not assume the ordinal characteristics of the severity estimation tasks. Most incorrect labels for ordinal labels likely occur between neighboring labels (such as Mayo 1 and 2) and unlikely between distant labels (such as Mayo 0 and 3). More formally, letting $y$ denote the true label, we can assume that the attached noisy label $\tilde{y}$ follows $\tilde{y}\sim\round{\mathcal{N}(y, \sigma^2)}$, where $\round{\cdot}$ is a round operation and $\sigma^2$ controls the noisiness of labels.

\par

To solve these two issues, we propose a novel framework called {\em self-relaxed joint training}, shown in Fig.~\ref{fig:overview}(b). Its key idea is that we have $y\sim\round{\mathcal{N}(\tilde{y}, \sigma^2)}$ from the above assumption; that is, given a noisy label $\tilde{y}$, we can estimate the distribution 
of the true label $y$. We can use this distribution as a {\em soft label} ``relaxed'' from the hard label $\tilde{y}$. Since this soft label is derived from the provided hard label, we refer to it as the ``{\em self}-relaxed'' soft label.
\par

Our new framework can efficiently solve the two issues. Specifically, it solves the second issue because it uses the soft labels derived under the above assumption about ordinal noisy labels. It is also possible to solve the first issue by using the loss from the soft label, $\mathcal{L}_\mathrm{s}$, instead of the loss by the hard label, $\mathcal{L}_\mathrm{h}$, for updating the models $f_1$ and $f_2$. Roughly speaking, using $\mathcal{L}_\mathrm{s}$ successfully weakens the negative impact of the selected samples with incorrect labels. Note that we still use $\mathcal{L}_\mathrm{h}$ for the criterion of the sample selection; our experimental results clearly show that this synergetic use of $\mathcal{L}_\mathrm{s}$ and  $\mathcal{L}_\mathrm{h}$ is significantly effective.
\par

We apply the proposed framework to the state-of-the-art methods in the traditional framework, that is, Co-teaching, CoDis, and JoCor, respectively, and evaluate how their performance is improved through the severity level classification task on two UC endoscopic image datasets and a retinal Diabetic Retinopathy (DR) dataset. The experimental results show significant improvements. Moreover, our methods (i.e., ``Co-teaching+Ours,'' ``CoDis+Ours,'' and ``JoCor+Ours'') also outperform state-of-the-art methods in various frameworks.
\par

The main contributions of this study are as follows:
\begin{itemize}
    \setlength{\itemsep}{0pt}   
    \item We propose a novel framework called self-relaxed joint training for learning with ordinal noisy labels. 
    \item The proposed framework introduces soft labels for representing ordinal noisy labels and a combination of hard and soft labels for sample selection.
    \item Our framework is versatile and thus can improve arbitrary methods in the traditional join-training framework, such as~\cite{co_teaching,jocor,codis}.
    \item Experimental evaluations on two UC datasets and the DR dataset with ordinal noisy labels show performance superiority over various state-of-the-art methods. 
\end{itemize}

\section{Related Work}\label{sec:related_work}

\subsection{Training a classifier with noisy labels} 

As noted in Section~\ref{sec:intro} and shown by Fig.~\ref{fig:overview}(a), the traditional joint-training framework to train a classifier with samples with noisy labels 
employ two techniques, sample selection and dual networks~\cite{noisy_label_survey}. The former selects small loss samples as samples with clean labels and discards large loss samples.
The latter uses two networks in a complementary manner. For example, Decoupling~\cite{decoupling} updates the networks only using samples where the predictions of two different networks differ. Co-teaching~\cite{co_teaching} aims to utilize the complementary performance of two networks by exchanging their clean samples. Co-teaching+~\cite{co_teaching_plus} combines Co-teaching and Decoupling for robust training. CoDis~\cite{codis} uses possibly clean samples that have high discrepancy prediction probabilities between two networks. On the other hand, JoCor~\cite{jocor} uses the same clean sample set for the two networks and introduces co-regularization to reduce divergence between the networks. \par
As shown by Fig.~\ref{fig:overview}(b), the proposed framework also uses sample selection and dual networks. Therefore, we can apply the proposed framework to the above methods, such as Co-teaching~\cite{co_teaching},  CoDis~\cite{codis}, and JoCor~\cite{jocor}. The later experiments will show that our methods (e.g., ``Co-teaching + Ours'') can improve their performance by employing the proposed framework with the self-relaxed soft labels.\par

In addition to the above methods in the traditional joint-training framework, various strategies have been proposed for handling noisy labels. For example, designing loss functions robust to noisy labels~\cite{NEURIPS2018_f2925f97,NEURIPS2019_8a1ee9f2,pmlr-v119-ma20c} is a popular strategy. Learning with label noise transition matrix uses the probability of label errors for loss functions~\cite{f_correction,jiang2022an,NEURIPS2020_5607fe88}. Learning to reweight examples uses importance weights for clean samples~\cite{reweihgt,pmlr-v80-ren18a,NEURIPS2020_8b9e7ab2}. Introducing various regularization strategies is also useful to suppress overfitting to the samples with incorrect labels ~\cite{mixup,cdr,Kim_2019_ICCV}. In the later experiment, we compare our methods with the most recent methods in the above strategies and prove that ours outperforms them. \par


\subsection{Ordinal regression with noisy labels}\  
Ordinal regression (, or ordinal classification) deals with classes with ordinal relationships and is well-studied in \cite{SORD,Li_2021_CVPR,Liu_2018_ECCV_Workshops,Niu_2016_CVPR,Shin_2022_CVPR}.
However, to the authors' best knowledge, only a few papers~\cite{garg, miccai_workshop_paper, unce_noisy} 
address ordinal regression with noisy labels. Among them, two papers~\cite{miccai_workshop_paper,unce_noisy}
rely on a strong assumption that multiple labels from multiple annotators are available for each sample. This strong assumption significantly narrows the scope of its applicability. In fact, as noted in Section~\ref{sec:intro}, finding multiple experts is often difficult, especially for rare diseases.\par
Consequently, only Garg et al.~\cite{garg} deals with ordinal regression with noisy labels under the same assumption as ours; only a single (noisy) label is attached to each sample. Their method decomposes a $C$-class ordinal regression problem into $(C-1)$ binary classification problems. Each binary problem aims to classify an input sample as belonging to a class higher or lower than class $c$. Each classifier uses a loss function that is robust to noisy labels. Consequently, Garg et al.~\cite{garg} uses a totally different 
methodology from ours. Moreover, the later experiment shows that ours largely outperforms their method.\par

\subsection{Label smoothing for classification task}\ 

Label smoothing (, or soft labeling) has been introduced to improve accuracy across many tasks~\cite{Szegedy_2016_CVPR,Zoph_2018_CVPR,Real_Aggarwal_Huang_Le_2019,chorowski2016towards,NIPS2017_3f5ee243}. In these studies, label smoothing is mainly used for regularization during training~\cite{label_smooth}. Moreover, previous studies have applied label smoothing to learning with noisy labels, demonstrating improvements in classification performance~\cite{pmlr-v119-lukasik20a,pmlr-v162-wei22b}. Additionally, for ordinal regression tasks, label-smoothing has been used~\cite{SORD}.  \par 

Our method is a novel framework to combine hard and soft labels for training with ordinal noisy labels. Furthermore, experimental results show that this combination improves classification accuracy compared to using soft labels alone. \par

\begin{algorithm}[t] 
1: {\bf Input:} Dataset $\tilde{\mathcal{D}}$, two networks $f_1$ and $f_2$ with initialized weights $\bm{\theta}_1$ and $\bm{\theta}_2$, learning rate $\eta$, noise rate $\epsilon$, epoch $T'$ and $T_{\max}$, iteration $t_{\max}$, temperature $\tau$;\\
\For{$T = 1,2,\dots,T_{\max}$}{
	
	2: {\bfseries Shuffle} training set $\tilde{\mathcal{D}}$;\\
	\For{$t = 1,\dots,t_{\max}$}
	{	
		3: {\bfseries Fetch} mini-batch $\tilde{\mathcal{B}}$ from $\tilde{\mathcal{D}}$;\\
            4: {\bfseries Select} clean samples from $\tilde{\mathcal{B}}$ by $\mathcal{L}_\mathrm{h}$ (with $\tau$);\\
            {\hphantom{hoge}$\mathcal{B}_1 \leftarrow \arg\min_{\mathcal{B}':|\mathcal{B}'|\ge R(T)|\tilde{\mathcal{B}}|}\mathcal{L}_\mathrm{h}(  f_1, \mathcal{B}')$; \hfill} \\
            {\hphantom{hoge}$\mathcal{B}_2 \leftarrow \arg\min_{\mathcal{B}':|\mathcal{B}'|\ge R(T)|\tilde{\mathcal{B}}|}\mathcal{L}_\mathrm{h}(  f_2, \mathcal{B}')$; \hfill} \\
            5: {\bfseries Derive} soft labels $\bm{l}_{\mathrm s}$ from $\bm{l}_{\mathrm h}$ for $\mathcal{B}_1, \mathcal{B}_2$ by Eq.(\ref{eq:soft_label});\\
		6: {\bfseries Update} networks; \\
            \hphantom{hoge}$\bm{\theta}_1 \leftarrow \bm{\theta}_1 - \eta\nabla \mathcal{L}_\mathrm{s}(f_1,\mathcal{B}_2)$; \hfill \\
		\hphantom{hoge}$\bm{\theta}_2 \leftarrow \bm{\theta}_2 - \eta\nabla \mathcal{L}_\mathrm{s}(f_2,\mathcal{B}_1)$; \hfill 
	}
	7: {\bfseries Update} $R(T) \leftarrow 1 - \min\left\lbrace  \frac{T}{T'} \epsilon, \epsilon \right\rbrace $;
}
8: {\bf Output:} two trained networks with $\bm{\theta}_1$ and $\bm{\theta}_2$.
\caption{Proposed framework with Co-teaching}
\label{alg:srjt}
\end{algorithm}

\section{Self-Relaxed Joint Training for Ordinal Noisy Labels}
\label{sec:method}

\subsection{Overview} 
\label{ssec:overview}

We propose a self-relaxed joint-training framework to train a classifier with ordinal noisy labels. As shown in Fig.~\ref{fig:overview}(b), for learning with ordinal noisy labels, the proposed framework uses  $\mathcal{L}_\mathrm{h}$ (the loss by hard labels) for the clean sample selection like the traditional methods, such as Co-teaching~\cite{co_teaching},  CoDis~\cite{codis}, and JoCor~\cite{jocor}. One main difference between our framework and the traditional framework is that ours also uses soft labels, which are relaxed versions of the hard labels, to update dual networks, $f_1$ and $f_2$. 
More specifically, we calculate $\mathcal{L}_\mathrm{s}$ (the loss by the soft label) and use it to update $f_1$ and $f_2$. Suppose a sample $\bm{x}$ is incorrectly labeled as 1 while its correct label is 2. If this sample passes the selection, $f_1$ and $f_2$ are updated to classify $\bm{x}$ as class 1 according to $\mathcal{L}_\mathrm{h}$. In contrast, using $\mathcal{L}_\mathrm{s}$ mitigates the negative impact of incorrectly labeled samples.
\par

As indicated by the similarity between Figs.~\ref{fig:overview}(a) and (b), the proposed framework can be applied to various methods with sample selection and dual networks. This application can be done by replacing $\mathcal{L}_\mathrm{h}$ for updating the models by $\mathcal{L}_\mathrm{s}$. This simple replacement, however, has a significant benefit on performance. 
In this section, we employ Co-teaching~\cite{co_teaching} as a backbone method because it is the most standard method with sample selection and dual networks. Algorithm~\ref{alg:srjt} shows the entire training procedure of ``Co-teaching + Ours,'' namely, the proposed framework applied to Co-teaching. The details of the algorithm will be explained below.\par
\subsection{Problem formulation} 

We assume an image dataset $\tilde{\mathcal{D}} = \{\bm{x}_i, \tilde{y}_i\}_{i=1}^{N}$, where $\bm{x}_i$ is the $i$-th image and $\tilde{y}_i\in \{1, 2,\ldots, C\}$ is its (noisy) class label and the $C$ classes have an ordinal relationship, $1 \prec 2 \prec \ldots \prec C$. For the case of four-level Mayo scores (Mayo 0, 1, 2, and 3) of UC images, the label $\tilde{y}_i=c$ means a UC image $\bm{x}_i$ is annotated with Mayo $(c-1)$.\par

Hereafter, we refer to the following one-hot vector $\bm{l}_{\mathrm h}(\tilde{y}_i)$ as a {\em hard label}:
\begin{equation}\label{eq:hard-label}
\bm{l}_{\mathrm h}(\tilde{y}_i) = (0,\ldots,\overset{\tilde{y}_i}{1}\ldots,0).
\end{equation}
Note again that $\tilde{y}_i$ is noisy due to the ambiguity of discrete severity levels; consequently, this hard label is also noisy.
\par

We use two deep neural networks $f_{\mathrm 1}(\bm{x}_i;\bm{\theta}_{\mathrm 1})$ and $f_{\mathrm 2}(\bm{x}_i;\bm{\theta}_{\mathrm 2})$, where $\bm{\theta}_{\mathrm 1}$ and $\bm{\theta}_{\mathrm 2}$ are weight parameters. The networks $f_{\mathrm 1}$ and $f_{\mathrm 2}$ are trained with soft labels in a complementary manner. We denote the outputs from $f_{n}$ ($n\in\{1, 2\}$) as $\bm{p}_n(\bm{x}_i) = (p_n^1(\bm{x}_i),\ldots, p_n^C(\bm{x}_i))$, where the element $p_n^c(\bm{x}_i)$ is the probability of class $c$ predicted at the softmax layer of $f_n$.
\par

\subsection{Soft labels from hard labels}
As noted in Section~\ref{sec:intro}, we can have the distribution of the clean class label $y_i$ from noisy class label $\tilde{y}_i$ by using the ordinal characteristics of the noisy labels. Specifically, if we can assume 
\begin{equation}
    \label{eq:ordinal-noise}
    \tilde{y}_i\sim\round{\mathcal{N}(y_i, \sigma^2)},
\end{equation}
we have $y_i\sim\round{\mathcal{N}(\tilde{y}_i, \sigma^2)}$. Considering this relationship and following \cite{SORD}, we define the {\em soft label} as $\bm{l}_{\mathrm s}(\tilde{y}_i) = (l_{\mathrm s}^1(\tilde{y}_i),\ldots, l_{\mathrm s}^C(\tilde{y}_i))$, where 
\begin{equation} 
    l_{\mathrm s}^c(\tilde{y}_i)= \frac{\exp(-|c - \tilde{y}_i|)}{\sum_{c'=1}^{C}\exp(-|c'-\tilde{y}_i|)}.
    \label{eq:soft_label}
\end{equation}
The soft label $\bm{l}_{\mathrm s}(\tilde{y}_i)$ has the largest value at $c=\tilde{y}_i$ and becomes smaller as the difference from $\tilde{y}_i$ becomes larger. Therefore, 
$\bm{l}_{\mathrm s}(\tilde{y}_i)$ is a relaxed version of the hard label $\bm{l}_{\mathrm h}(\tilde{y}_i)$ of Eq.(\ref{eq:hard-label})

\subsection{Sample selection by hard labels}
\label{ssec:sample_selection}
A sample $\bm{x}_i$ is selected as clean when its $\mathcal{L}_{\mathrm h}$ is small:
\begin{eqnarray}
    \mathcal{L}_{\mathrm h}(f_n,\tilde{\mathcal{B}}) &=& -\sum_{\{\bm{x}_i,\tilde{y}_i\} \in \tilde{\mathcal{B}}}\sum_{c=1}^{C}l_{\mathrm h}^c(\tilde{y}_i)\log{p_n^c(\bm{x}_i)}\nonumber \\ 
    &=& -\sum_{\{\bm{x}_i,\tilde{y}_i\} \in \tilde{\mathcal{B}}}\log{p_n^{\tilde{y}_i}(\bm{x}_i)}.
       \label{eq:h_CE}
\end{eqnarray}
Specifically, among the samples in a mini-batch $\tilde{\mathcal{B}}$, the $R(T)|\tilde{\mathcal{B}}|$ samples with the smallest loss value are selected at the $T$th epoch, where $R(T)\in (0,1]$ is the selection rate.\par
Precisely speaking, the softmax layer with a temperature $\tau \in (0,1)$ is used to make $\bm{p}_n(\bm{x}_i)$ more peaky for sample selection. (This is a practical remedy for the over-smoothed output probabilities by the network trained with soft labels.)
Moreover, following the traditional setting~\cite{co_teaching,jocor,codis}, we used $R(T)=1-\mathrm{min} \{\frac{T}{T'}\epsilon, \epsilon\}$, where $T'$ is a hyperparameter and $\epsilon$ is the noise rate (i.e., the ratio of incorrect labels) of the dataset. Under this setting, 
more samples than $(1-\epsilon)|\tilde{\mathcal{B}}|$ are selected at earlier epochs $T<T'$; this treatment is helpful to avoid the memorization effect~\cite{memorization,zhang2017understanding}, which is a phenomenon that networks tend to overfit (very) clean samples. At later epochs $T\geq T'$, $(1-\epsilon)|\tilde{\mathcal{B}}|$ samples are selected.

\subsection{Updating the network parameters by soft labels} 
For updating the networks, we use the loss function $\mathcal{L}_{\mathrm s}$ evaluated with soft labels $\bm{l}_{\mathrm s}$. In addition, each network uses the samples selected by the other network for its updating, following \cite{co_teaching,codis}. Consequently, the network $f_1$ is updated with the loss function:
\begin{equation}
    \label{eq:s_CE}
    \mathcal{L}_{\mathrm s}(f_1,\mathcal{B}_2) = -\sum_{\{x_i,\tilde{y}_i\} \in \mathcal{B}_2}\sum_{c=1}^{C}l_{\mathrm s}^c(\tilde{y}_i)\log{p_n^c(\bm{x}_i)},
\end{equation}
where $\mathcal{B}_2$ is a mini-batch with the samples selected by $f_2$. For $f_2$, we use $\mathcal{L}_{\mathrm s}(f_2,\mathcal{B}_1)$. 
\par

\subsection{How to apply our framework to other methods}
\label{ssec:plugin}

As noted in Section~\ref{ssec:overview}, our framework can be applied to any method in the joint-training framework of Fig.~\ref{fig:overview}(a). The application is done with two modifications
of the traditional framework. First, as shown in Algorithm~\ref{alg:srjt} of ``Co-teaching + Ours,'' ours employs $\mathcal{L}_{\mathrm s}$ for updating the networks, whereas the original Co-teaching~\cite{co_teaching} uses $\mathcal{L}_{\mathrm h}$. Second, the temperature $\tau$ is introduced to the softmax layer of the networks. Similar modifications of CoDis~\cite{codis} and JoCor~\cite{jocor} give ``CoDis+Ours'' and ``JoCor+Ours.'' (See the supplementary file for their details.) These modifications seem simple but result in a significant improvement in sample selection and classification.\par

\begin{table*}[t]
\renewcommand{\arraystretch}{0.9}
\small
\centering
\caption{Classification results on \underline{LIMUC} with \underline{Quasi-Gaussian noise}. Following tradition, the test accuracy (Acc.), mean absolute error (MAE), and macro F1 (mF1)are averaged over the last ten epochs. The mean and standard deviations of five-fold cross-validation are shown. The best and second-best results are highlighted in \textcolor{red}{red}\ and  \textcolor{blue}{{blue}}, respectively. For plugin settings, improved results are shown by {\bf bold}.}
\begin{tabular}{lcccccc}
\toprule
\multirow{2}{*}{Method} & \multicolumn{3}{c}{Noise rate: $\epsilon=0.2$} & \multicolumn{3}{c}{Noise rate: $\epsilon=0.4$} \\
\cmidrule(lr){2-4} \cmidrule(lr){5-7} & Acc.$\uparrow$  & MAE$\downarrow$ & mF1$\uparrow$ & Acc.$\uparrow$ & MAE$\downarrow$ & mF1$\uparrow$ \\
\midrule
Standard            & 0.673$\pm$0.013 & 0.383$\pm$0.020 & 0.572$\pm$0.011 & 0.556$\pm$0.017 & 0.564$\pm$0.025 & 0.463$\pm$0.019 \\
Sord~\cite{SORD}                & 0.701$\pm$0.018 & 0.320$\pm$0.020 & 0.621$\pm$0.020 & 0.582$\pm$0.017 & 0.469$\pm$0.016 & 0.518$\pm$0.021 \\
Label-smooth~\cite{label_smooth} & 0.684$\pm$0.021 & 0.358$\pm$0.029 & 0.582$\pm$0.023 & 0.600$\pm$0.008 & 0.477$\pm$0.009 & 0.498$\pm$0.012 \\
F-correction~\cite{f_correction}       & 0.666$\pm$0.020 & 0.389$\pm$0.024 & 0.562$\pm$0.012 & 0.562$\pm$0.016 & 0.562$\pm$0.027 & 0.469$\pm$0.018 \\
Reweight~\cite{reweihgt}           & 0.670$\pm$0.012 & 0.387$\pm$0.016 & 0.567$\pm$0.009 & 0.556$\pm$0.013 & 0.560$\pm$0.017 & 0.465$\pm$0.010 \\
Mixup~\cite{mixup}              & 0.676$\pm$0.016 & 0.374$\pm$0.026 & 0.578$\pm$0.015 & 0.596$\pm$0.012 & 0.494$\pm$0.012 & 0.486$\pm$0.012 \\
CDR~\cite{cdr}                 & 0.664$\pm$0.016 & 0.399$\pm$0.024 & 0.560$\pm$0.013 & 0.550$\pm$0.005 & 0.585$\pm$0.009 & 0.454$\pm$0.010 \\
Garg~\cite{garg}               & 0.681$\pm$0.012 & 0.377$\pm$0.015 & 0.497$\pm$0.018 & 0.600$\pm$0.019 & 0.579$\pm$0.026 & 0.381$\pm$0.010 \\
\midrule
Co-teaching~\cite{co_teaching}                             & 0.699$\pm$0.015 & 0.336$\pm$0.016 & 0.602$\pm$0.010 & 0.647$\pm$0.009 & 0.399$\pm$0.009 & 0.550$\pm$0.018 \\
\rowcolor{gray!15}Co-teaching + Ours    & {\bf 0.726$\pm$0.010} & {\bf 0.292$\pm$0.012} & {\bf 0.640$\pm$0.011}& {\color{blue}\bf 0.689$\pm$0.012} & {\color{blue}\bf0.344$\pm$0.016} & {\color{blue}\bf0.575$\pm$0.023} \\
\midrule
JoCor~\cite{jocor}                                   & 0.718$\pm$0.010 & 0.310$\pm$0.011 & 0.624$\pm$0.005 & 0.679$\pm$0.016 & 0.363$\pm$0.019 & 0.536$\pm$0.077 \\
\rowcolor{gray!15} JoCor + Ours         & {\color{red}\bf0.733$\pm$0.007} & {\color{red}\bf0.286$\pm$0.008} & {\color{red}\bf0.644$\pm$0.010} & {\color{red}\bf0.702$\pm$0.007} & {\color{red}\bf0.328$\pm$0.012} & {\color{red}\bf0.595$\pm$0.023} \\
\midrule
CoDis~\cite{codis}                                   & 0.695$\pm$0.012 & 0.342$\pm$0.015 & 0.600$\pm$0.009 & 0.619$\pm$0.009 & 0.440$\pm$0.011 & 0.526$\pm$0.014 \\
\rowcolor{gray!15}CoDis + Ours          & {\color{blue}\bf0.727$\pm$0.008} & {\color{blue}\bf0.290$\pm$0.009} & {\color{blue}\bf0.642$\pm$0.010} & {\bf 0.673$\pm$0.005} & {\bf 0.360$\pm$0.005} & {\bf 0.573$\pm$0.012} \\
\bottomrule
\label{tab:results_limuc_inverse}
\end{tabular}
\end{table*}

\begin{table*}[t]
\renewcommand{\arraystretch}{0.9}
\small
\centering
\caption{Classification results on our \underline{private UC dataset} with \underline{Quasi-Gaussian noise}. See the caption of Table~\ref{tab:results_limuc_inverse} for details.}
\begin{tabular}{lcccccc}
\toprule
\multirow{2}{*}{Method} & \multicolumn{3}{c}{Noise rate: $\epsilon=0.2$} & \multicolumn{3}{c}{Noise rate: $\epsilon=0.4$} \\
\cmidrule(lr){2-4} \cmidrule(lr){5-7} & Acc.$\uparrow$  & MAE$\downarrow$ & mF1$\uparrow$ & Acc.$\uparrow$ & MAE$\downarrow$ & mF1$\uparrow$ \\
\midrule
Standard            & 0.756$\pm$0.009 & 0.291$\pm$0.008 & 0.530$\pm$0.015 & 0.640$\pm$0.022 & 0.474$\pm$0.023 & 0.436$\pm$0.016 \\
Sord~\cite{SORD}               & 0.792$\pm$0.007 & 0.226$\pm$0.007 & 0.579$\pm$0.033 & 0.678$\pm$0.018 & 0.358$\pm$0.023 & 0.513$\pm$0.022 \\
Label-smooth~\cite{label_smooth} & 0.783$\pm$0.018 & 0.250$\pm$0.020 & 0.564$\pm$0.038 & 0.694$\pm$0.019 & 0.367$\pm$0.021 & 0.485$\pm$0.022 \\
F-correction~\cite{f_correction}       & 0.755$\pm$0.014 & 0.293$\pm$0.017 & 0.545$\pm$0.033 & 0.635$\pm$0.021 & 0.472$\pm$0.032 & 0.447$\pm$0.019 \\
Reweight~\cite{reweihgt}           & 0.762$\pm$0.016 & 0.286$\pm$0.021 & 0.549$\pm$0.009 & 0.634$\pm$0.023 & 0.484$\pm$0.043 & 0.434$\pm$0.017 \\
Mixup~\cite{mixup}              & 0.765$\pm$0.011 & 0.281$\pm$0.015 & 0.530$\pm$0.026 & 0.673$\pm$0.014 & 0.409$\pm$0.013 & 0.450$\pm$0.014 \\
CDR~\cite{cdr}                 & 0.752$\pm$0.004 & 0.295$\pm$0.007 & 0.535$\pm$0.006 & 0.631$\pm$0.017 & 0.488$\pm$0.024 & 0.443$\pm$0.019 \\
Garg~\cite{garg}               & 0.734$\pm$0.020 & 0.311$\pm$0.022 & 0.467$\pm$0.023 & 0.664$\pm$0.031 & 0.472$\pm$0.051 & 0.342$\pm$0.015 \\
\midrule
Co-teaching~\cite{co_teaching}                             & 0.790$\pm$0.009 & 0.237$\pm$0.008 & 0.594$\pm$0.014 & 0.719$\pm$0.015 & 0.324$\pm$0.016 & 0.512$\pm$0.036 \\
\rowcolor{gray!15}Co-teaching + Ours    & {\color{blue}\bf0.818$\pm$0.008} & {\color{blue}\bf0.200$\pm$0.009} & {\color{blue}\bf0.599$\pm$0.031} & {\bf 0.776$\pm$0.015} & {\bf 0.256$\pm$0.019} & {\color{blue}\bf0.555$\pm$0.031} \\
\midrule
JoCor~\cite{jocor}                                   & 0.812$\pm$0.005 & 0.209$\pm$0.008 & {\color{red}0.601$\pm$0.018} & {\color{red}0.787$\pm$0.006} & {\color{red}0.242$\pm$0.005} & 0.547$\pm$0.047 \\
\rowcolor{gray!15} JoCor + Ours         & {\color{red}\bf0.819$\pm$0.006} & {\color{red}\bf0.197$\pm$0.005} & 0.585$\pm$0.031 & {\color{blue}0.785$\pm$0.011} & {\color{blue}0.242$\pm$0.015} & {\color{red}\bf0.558$\pm$0.040} \\
\midrule
CoDis~\cite{codis}                                   & 0.784$\pm$0.010 & 0.246$\pm$0.010 & 0.585$\pm$0.015 & 0.699$\pm$0.022 & 0.350$\pm$0.025 & 0.508$\pm$0.027 \\
\rowcolor{gray!15}CoDis + Ours          & {\bf 0.808$\pm$0.007} & {\bf 0.209$\pm$0.007} & {\bf 0.598$\pm$0.019} & {\bf 0.756$\pm$0.016} & {\bf 0.271$\pm$0.015} & {\bf 0.549$\pm$0.034} \\
\bottomrule
\label{tab:results_private_inverse}
\end{tabular}
\vspace{-5pt}
\end{table*}

\begin{table*}[t]
\renewcommand{\arraystretch}{0.9}
\small
\centering
\caption{Classification results on our \underline{private UC dataset} with \underline{Truncated-Gaussian noise}.  See the caption of Table~\ref{tab:results_limuc_inverse} for details.}
\begin{tabular}{lcccccc}
\toprule
\multirow{2}{*}{Method} & \multicolumn{3}{c}{Noise rate: $\epsilon=0.2$} & \multicolumn{3}{c}{Noise rate: $\epsilon=0.4$} \\
\cmidrule(lr){2-4} \cmidrule(lr){5-7} & Acc.$\uparrow$  & MAE$\downarrow$ & mF1$\uparrow$ & Acc.$\uparrow$ & MAE$\downarrow$ & mF1$\uparrow$ \\
\midrule
Standard            & 0.760$\pm$0.010 & 0.270$\pm$0.010 & 0.559$\pm$0.012 & 0.629$\pm$0.028 & 0.418$\pm$0.027 & 0.423$\pm$0.038 \\
Sord~\cite{SORD}                & 0.792$\pm$0.009 & 0.225$\pm$0.008 & 0.594$\pm$0.026 & 0.680$\pm$0.011 & 0.340$\pm$0.009 & 0.485$\pm$0.025 \\
Label-smooth~\cite{label_smooth} & 0.786$\pm$0.008 & 0.243$\pm$0.008 & 0.590$\pm$0.040 & 0.665$\pm$0.022 & 0.375$\pm$0.018 & 0.448$\pm$0.019 \\
F-correction~\cite{f_correction}       & 0.763$\pm$0.008 & 0.263$\pm$0.007 & 0.563$\pm$0.035 & 0.655$\pm$0.023 & 0.379$\pm$0.024 & 0.456$\pm$0.030 \\
Reweight~\cite{reweihgt}           & 0.761$\pm$0.019 & 0.271$\pm$0.018 & 0.559$\pm$0.022 & 0.610$\pm$0.017 & 0.436$\pm$0.016 & 0.431$\pm$0.011 \\
Mixup~\cite{mixup}              & 0.757$\pm$0.016 & 0.273$\pm$0.013 & 0.551$\pm$0.019 & 0.646$\pm$0.022 & 0.398$\pm$0.020 & 0.445$\pm$0.022 \\
CDR~\cite{cdr}                 & 0.758$\pm$0.014 & 0.272$\pm$0.012 & 0.574$\pm$0.015 & 0.620$\pm$0.017 & 0.430$\pm$0.012 & 0.412$\pm$0.019 \\
Garg~\cite{garg}               & 0.737$\pm$0.019 & 0.303$\pm$0.017 & 0.481$\pm$0.013 & 0.610$\pm$0.036 & 0.625$\pm$0.061 & 0.233$\pm$0.010 \\
\midrule
Co-teaching~\cite{co_teaching}                             & 0.788$\pm$0.009 & 0.236$\pm$0.008 & 0.599$\pm$0.031 & 0.702$\pm$0.022 & 0.328$\pm$0.020 & 0.490$\pm$0.036 \\
\rowcolor{gray!15}Co-teaching + Ours    & {\color{blue}\bf0.815$\pm$0.010} & {\color{blue}\bf0.202$\pm$0.008} & {\color{blue}\bf0.621$\pm$0.035} & {\bf 0.748$\pm$0.031} & {\bf 0.282$\pm$0.033} & {\bf0.491$\pm$0.032} \\
\midrule
JoCor~\cite{jocor}                                   & 0.812$\pm$0.011 & 0.209$\pm$0.011 & {\color{red}0.621$\pm$0.032} & 0.744$\pm$0.032 & 0.286$\pm$0.031 & 0.474$\pm$0.028 \\
\rowcolor{gray!15} JoCor + Ours         & {\color{red}\bf0.817$\pm$0.007} & {\color{red}\bf0.199$\pm$0.007} & 0.607$\pm$0.040 & {\bf\color{red}0.760$\pm$0.027} & {\color{red}\bf0.266$\pm$0.025} & {\color{blue}\bf0.493$\pm$0.014} \\
\midrule
CoDis~\cite{codis}                                   & 0.780$\pm$0.011 & 0.246$\pm$0.010 & 0.585$\pm$0.023 & 0.675$\pm$0.015 & 0.357$\pm$0.013 & 0.483$\pm$0.025 \\
\rowcolor{gray!15}CoDis + Ours          & {\bf 0.809$\pm$0.009} & {\bf 0.207$\pm$0.007} & {\bf 0.611$\pm$0.038} & {\color{blue}\bf 0.749$\pm$0.024} & {\color{blue}\bf 0.272$\pm$0.023} & {\color{red}\bf 0.531$\pm$0.028} \\
\bottomrule
\label{tab:results_private_neighbor}
\end{tabular}
\vspace{-5pt}
\end{table*}

\section{Experimental Evaluation}\label{sec:experiment}
\subsection{Experimental Setup\label{sec:setup}}
\noindent{\bf Datasets with clean labels.}\ 
We used two datasets of UC endoscopic images. One is the publicly available dataset, LIMUC~\cite{limuc}, which contains 11,276 images from 564 patients. The other is our private dataset from 388 patients at Kyoto Second Red Cross Hospital. For both datasets, three medical experts attached the Mayo scores to ensure clean labels at the cost of annotation effort. The distribution of labels (Mayo 0, 1, 2, and 3) is 6,105, 3,052, 1,254, and 865 images on LIMUC and 6,678, 1,995, 1,395, and 197 images on the private dataset, respectively. All images in the two datasets were resized to 256 $\times$ 256 pixels.\par

\noindent{\bf Datasets with noisy labels.}\ 
We created noisy-labeled datasets from the clean-labeled datasets with label perturbations for the quantitative evaluation. The perturbations are specified by the label transition matrix $P=[P_{ij}]$, where the $P_{ij} = \mathrm{Pr}(\tilde{y}=j|y=i)$ (i,e., probability of mistaking class $i$ to class $j$). 
We employed two types of $P$. The first type is {\bf Quasi-Gaussian}, which is an approximated version of the Gaussian perturbation of Eq.~(\ref{eq:ordinal-noise}). Specifically, by following Garg et al.~\cite{garg}, we set $P_{ij}=\frac{\rho}{|i-j|}, \forall i\ne j$ and $P_{ii}=1-\sum_{j\backslash i}P_{ij},\ \forall i$. Here, $\rho$ is a parameter for noise strength. The second type is {\bf Truncated-Gaussian}, which mimics the case that experts do not make severe mistakes. Specifically, $P_{ij} = 1 - \rho$ for $|i-j|=1$ and $P_{ij} = 0$ for $|i-j|>1$. This means, for example, that a UC image with Mayo 1 is only mislabeled as Mayo 0 or 2 and not 3. The diagonal element $P_{ii}$ is calculated by the same equation as Quasi-Gaussian.
\par

\noindent{\bf Noise rate $\epsilon$.}\ 
The noise rate $\epsilon$ indicates the ratio of samples with incorrect labels in the dataset. According to the past attempts with noisy labels~\cite{co_teaching,jocor,codis,co_teaching_plus}, we use two rates, i.e., $\epsilon=0.2$ (moderately noisy) and $0.4$ (extremely noisy). For setting $\epsilon$ at $0.2$ and $0.4$, we specify $\rho$ to appropriate values. For Quasi-Gaussian,  we set $\rho$ at $0.1$ and $0.2$ to have $\epsilon=0.2$ and $0.4$, respectively. Similarly, we set $\rho$ at $0.15$ and $0.3$ for Truncated-Gaussian, respectively.

\noindent{\bf Real-world noisy dataset.}\ We used the Diabetic Retinopathy~(DR) dataset\footnote{https://kaggle.com/competitions/diabetic-retinopathy-detection} as a real-world noisy dataset, where noisy labels naturally present. This dataset includes 35,108 retinal images, classified into five ordinal levels of DR severity: No DR (lowest), Mild DR, Moderate DR, Severe DR, and Proliferative DR (highest). The sample sizes for these levels are 25,802, 2,438, 5,288, 872, and 708, respectively. For training the proposed method, we set the noise rate $\epsilon = 0.3$, as the DR dataset is estimated to contain approximately 30\% noisy ordinal labels due to through annotation~\cite{unce_noisy}. All images were resized to 256 $\times$ 256 pixels.\par

\noindent{\bf Comparative methods.}\ We compare the proposed method with the traditional and state-of-art methods for learning with noisy labels: Co-teaching~\cite{co_teaching}, JoCor~\cite{jocor}, and CoDis~\cite{codis}. 
In addition, we employ the recent methods that use robust loss functions, reweighting examples, and regularizations: F-correction~\cite{f_correction}, Reweight~\cite{reweihgt}, Mixup~\cite{mixup}, and CDR~\cite{cdr}. Moreover, we use Garg et al.~\cite{garg}, the only ordinal regression method with noisy labels. Furthermore, we employ a label smoothing method (Label-smooth)~\cite{label_smooth,Szegedy_2016_CVPR}. 
As two simple baselines, we use Standard and Sord~\cite{SORD}. Both are just a single CNN without any sample selection, and the former is trained with the hard labels, whereas the latter is trained with the soft labels.
\par

As noted in Section~\ref{ssec:plugin}, we applied our framework to Co-teachig~\cite{co_teaching}, JoCor~\cite{jocor}, and CoDis~\cite{codis}. 
We refer to the resulting method as ``Co-teaching + Ours,''  ``JoCor + Ours,'' and  ``CoDis + Ours.''  \par

\noindent{\bf Evaluation metrics.}\ 
We used accuracy and mean absolute error (MAE) for the test set to evaluate the classification performance. MAE is computed as the mean of the absolute differences between the predicted class labels and the true (and clean) class label. In addition, macro-F1 was also evaluated for evaluation metrics because of class imbalance in the two UC datasets and the DR dataset. Following the tradition of the papers on learning with noisy labels~\cite{co_teaching,jocor,codis,co_teaching_plus}, all metrics were calculated as averages over the last ten epochs.
\par

We used five-fold cross-validation in all evaluations. Each clean-label dataset was split into training, validation, and test sets with 60, 20, and 20\%, respectively, by random patient-disjoint sampling. Then, for two UC datasets, as noted above, we made the training and validation sets noisy by the label transition matrix $P$ to have $\epsilon=0.2$ or $0.4$. This means we use the noisy validation set to simulate a practical situation where no clean dataset is available. Only the test set is clean for the meaningful evaluation. The methods with a dual-network architecture have two networks with slightly different accuracies. In the following results, we show the average of the performance of two networks.\par

\begin{figure}[t]
    \centering
    \includegraphics[width=\linewidth]{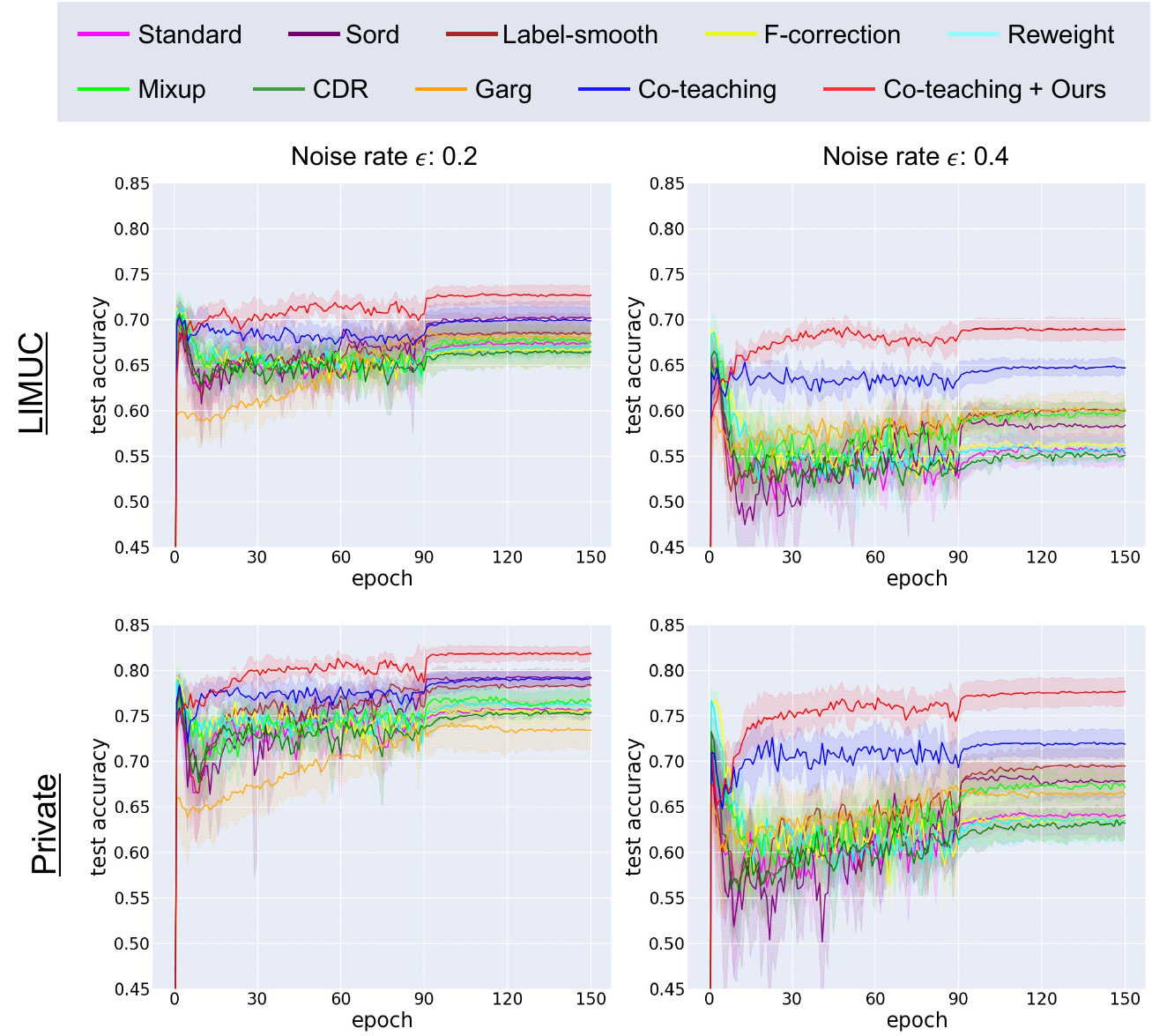}
    \caption{Test accuracy curves. 
    The width of the shading indicates the standard deviation in cross-validation.}
    \label{fig:test_acc}
    \vspace{-5pt}
\end{figure}

\noindent{\bf Implementation details.}\ 
We used ResNet-18 pretrained on ImageNet for all methods. We also used Adam as the optimizer and set the batch size to 64 for two UC datasets and 128 for the DR dataset. Training epochs were 150 in total, and an initial learning rate was $1\times10^{-4}$. The softmax temperature $\tau$ was set at 0.1. Following the traditional methods~\cite{co_teaching,jocor,codis}, $T'$ was set at $5$. For fair comparisons, L2 regularization of parameters and a multi-step learning rate scheduler were used to avoid overfitting noisy labels. We used random horizontal flipping and random cropping (224 $\times$ 224) for data augmentation in training. (Note that center-cropped 224 $\times$ 224 images were 
fed to the networks in the evaluation step.)

\subsection{Quantitative Evaluation Results}
\label{ssec:results}

Tables~\ref{tab:results_limuc_inverse} and \ref{tab:results_private_inverse} show the quantitative evaluation results for LIMUC and our private UC dataset, respectively, where Quasi-Gaussian was used to make the datasets noisy. The proposed methods (``$*$ + Ours'') achieve the best or second-best results across all metrics in both datasets and at both noise rates. We can also confirm that our self-relaxed joint-training framework boosts the performance of traditional methods, i.e., Co-teaching, JoCor, and CoDis. Moreover, ours outperforms Garg et al.~\cite{garg}; one possible reason is they do not assume Gaussian-like characteristics of ordinal noisy labels. From those results, we can conclude that introducing soft labels representing the distribution of true labels is useful for dealing with ordinal noisy labels. 
\par 
Table~\ref{tab:results_private_neighbor} shows the results on our private dataset under Truncated-Gaussian noise. Our methods (``$*$+ Ours'') still outperform the other methods like Table~\ref{tab:results_private_inverse}. Compared to Table~\ref{tab:results_private_inverse}, the individual accuracies in Table~\ref{tab:results_private_neighbor} are slightly lower. Under the same noise rate $\epsilon$, Truncated-Gaussian shows slightly inferior results than Quasi-Gaussian. This is because Quasi-Gaussian contains ``obviously incorrect'' labels (such as the ``Mayo 0'' label for a ``Mayo 2'' sample) to some extent, and they are easily discarded. In contrast, Truncated-Gaussian contains only ambiguous samples with only one-level errors, and therefore, it was difficult for all methods to discard the samples.

\begin{figure}[t]
    \centering
    \includegraphics[width=\linewidth]{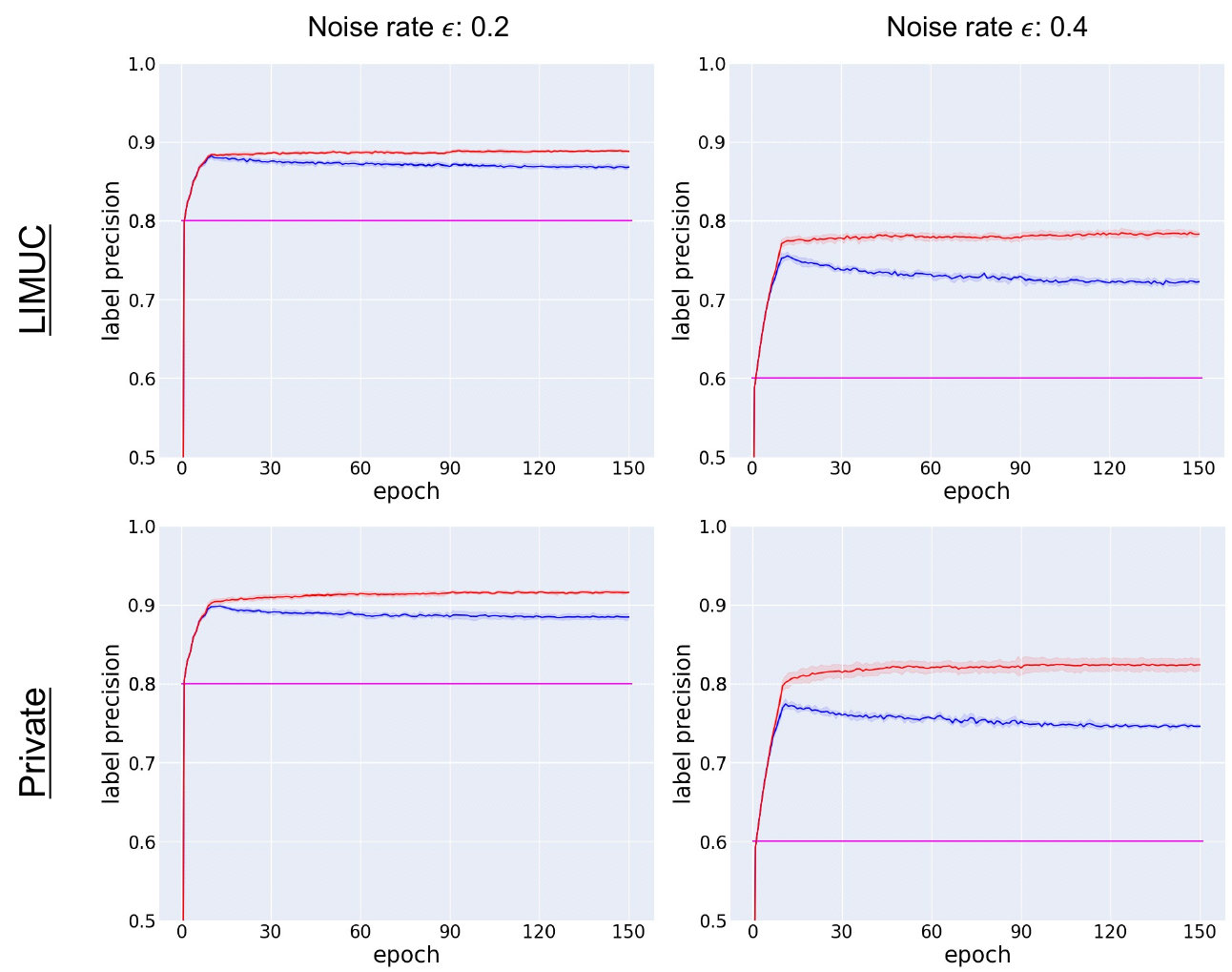}
    \caption{Label precision curves. The \blue{blue} and \red{red} curves show the label precisions by ``Co-teaching'' and ``Co-teaching + Ours,'' respectively. The pink horizontal line shows $(1-\epsilon)$.}
    \label{fig:label_precision}
    \vspace{-5pt}
\end{figure}

\subsection{Test accuracy curves}
Fig.~\ref{fig:test_acc} shows the test accuracy curves for the individual methods on two UC datasets with Quasi-Gaussian noise. The comparative methods show a sharp increase in their test accuracy in very early epochs ($<5$) by using a limited number of ``very'' clean samples to train their model. Then, the comparative models often start ``memorizing'' the samples with incorrect labels. The training accuracy is further improved through memorization, but the test accuracy degrades. This phenomenon is known as the memorization effect~\cite{co_teaching} in learning with noisy labels. In contrast, our method (``Co-teaching + Ours'') could avoid the memorization effect, and its test accuracy is roughly improved with the training steps. As shown in Section~\ref{ssec:lp}, our method could select clean samples much more than Co-teaching, even under the higher noise rate condition, and therefore, avoid the memorization effect more successfully than the other methods.

\begin{table*}[t]
\renewcommand{\arraystretch}{0.9}
\small
\centering
\caption{Results of \underline{LIMUC dataset} under different loss usages for sample selection and updating. The best and second-best results are highlighted in \textcolor{red}{red}\ and  \textcolor{blue}{{blue}}, respectively. \label{tab:abl-limuc}}
\begin{tabular}{cccccccc}
\toprule
\multirow{2}{*}{Selection} & \multirow{2}{*}{Updating} &\multicolumn{3}{c}{Noise rate: $\epsilon=0.2$} & \multicolumn{3}{c}{Noise rate: $\epsilon=0.4$} \\
\cmidrule(lr){3-5} \cmidrule(lr){6-8} & & Acc.$\uparrow$  & MAE$\downarrow$ & mF1$\uparrow$ & Acc.$\uparrow$ & MAE$\downarrow$ & mF1$\uparrow$ \\
\midrule
hard & hard & 0.699$\pm$0.015 & 0.336$\pm$0.016 & 0.602$\pm$0.010 & 0.647$\pm$0.009 & 0.399$\pm$0.009 & {\color{blue}0.550$\pm$0.018} \\
soft & soft & {\color{red}0.726$\pm$0.010} & {\color{blue}0.293$\pm$0.011} & {\color{blue}0.633$\pm$0.014} & {\color{blue}0.677$\pm$0.011} & {\color{blue}0.367$\pm$0.013} & 0.530$\pm$0.009 \\
\rowcolor{gray!15}hard & soft & {\color{red}0.726$\pm$0.010} & {\color{red}0.292$\pm$0.012} & {\color{red}0.640$\pm$0.011} & {\color{red}0.689$\pm$0.012} & {\color{red}0.344$\pm$0.016} & {\color{red}0.575$\pm$0.023} \\
\bottomrule
\end{tabular}
\bigskip\\
\renewcommand{\arraystretch}{0.9}
\small
\centering
\caption{Results of \underline{our private dataset} under different loss usages for sample selection and updating.\label{tab:abl-private}}
\begin{tabular}{cccccccc}
\toprule
\multirow{2}{*}{Selection} & \multirow{2}{*}{Updating} &\multicolumn{3}{c}{Noise rate: $\epsilon=0.2$} & \multicolumn{3}{c}{Noise rate: $\epsilon=0.4$} \\
\cmidrule(lr){3-5} \cmidrule(lr){6-8} & & Acc.$\uparrow$  & MAE$\downarrow$ & mF1$\uparrow$ & Acc.$\uparrow$ & MAE$\downarrow$ & mF1$\uparrow$ \\
\midrule
hard & hard & 0.790$\pm$0.009 & 0.237$\pm$0.008 & 0.594$\pm$0.014 & 0.719$\pm$0.015 & 0.324$\pm$0.016 & 0.512$\pm$0.036 \\
soft & soft & {\color{blue}0.812$\pm$0.011} & {\color{blue}0.204$\pm$0.012} & {\color{red}0.602$\pm$0.017} & {\color{blue}0.757$\pm$0.023} & {\color{blue}0.276$\pm$0.025} & {\color{blue}0.529$\pm$0.036} \\
\rowcolor{gray!15}hard & soft      & {\color{red}0.818$\pm$0.008} & {\color{red}0.200$\pm$0.009} & {\color{blue}0.599$\pm$0.031} & {\color{red}0.776$\pm$0.015} & {\color{red}0.256$\pm$0.019} & {\color{red}0.555$\pm$0.036} \\
\bottomrule
\end{tabular}
\end{table*}

\subsection{Label precision curve\label{ssec:lp}}

Fig.~\ref{fig:label_precision} shows, on UC datasets with Quasi-Gaussian noise, the change in label precision, which evaluates the ratio of clean samples among all the selected samples. The backbone method is Co-teaching. The pink horizontal lines ($1-\epsilon$) indicate the label precision under random sample selection. The proposed method (``Co-teaching + Ours,''  the red curve) shows far better label precisions than random selection. In addition, we can see that the selection performance is stable and thus does not degrade with epochs. In contrast, Co-teaching (the blue curve) shows the degradation with epochs because the network updated by hard labels is not reliable enough for sample selection.

\subsection{How good is our combination of hard and soft labels?}
\label{ssec:ablation}
As shown in Fig.~\ref{fig:overview}(b), our method uses the hard labels $\bm{l}_\mathrm{h}$ and their corresponding loss $\mathcal{L}_\mathrm{h}$ for sample selection and the soft labels $\bm{l}_\mathrm{s}$ and their loss $\mathcal{L}_\mathrm{s}$ for updating the model parameters. Tables~\ref{tab:abl-limuc} and \ref{tab:abl-private} show how this combination is appropriate for learning with ordinal (Quasi-Gaussian) noisy labels of LIMUC and the private dataset, respectively. In these experiments, Co-teaching is used as the backbone method; therefore, the upper-most ``hard \& hard'' case corresponds to the original Co-teaching~\cite{co_teaching} and the lower-most ``hard \& soft'' to ``Co-teaching + Ours.'' \par
The most important comparison with these tables is ``hard \& soft'' (i.e., ours) vs ``soft \& soft.'' One might think that the consistent `` soft \& soft'' is more reasonable than our inconsistent ``hard \& soft.'' However, this is not true. Using soft labels for sample selection is less suitable than using hard labels, especially for the private dataset.
\par

\subsection{Effectiveness in performance on a real noisy dataset}
\label{ssec:real-world}
Table~\ref{tab:results_dr} shows the classification results on the DR dataset, a real-world noisy dataset. Our methods (``$*$ + Ours'') achieve either the best or second-best results across all metrics. Additionally, the accuracy and MAE of traditional joint-training methods (i.e., Co-teaching, JoCor, and CoDis) are significantly worse than those of comparison methods. This is because traditional methods were designed for standard noisy labels. In contrast, our self-relaxed joint training framework boosts performance as our framework introduces modifications that allow these methods to handle ordinal noisy labels. From these results, we can conclude that the proposed method is effective for practical medical diagnosis tasks that contain ordinal noisy labels. 
\par

\begin{table}[t]
\renewcommand{\arraystretch}{0.9}
\small
\centering
\caption{Classification results on the DR dataset. The mean of five-fold cross-validation is shown. The best and second-best results are highlighted in \textcolor{red}{red}\ and  \textcolor{blue}{{blue}}, respectively. For plugin settings, improved results are shown by {\bf bold}.}
\begin{tabular}{lccc}
\toprule
Method & Acc.$\uparrow$ & MAE$\downarrow$ & mF1$\uparrow$ \\
\midrule
Standard            & 0.731 & 0.463 & 0.356 \\
Sord~\cite{SORD}                & {\color{blue}0.744} & {\color{blue}0.428} & 0.373  \\
Label-smooth~\cite{label_smooth}                & 0.742 & 0.445 & 0.356  \\
F-correction~\cite{f_correction}       & 0.729 & 0.457 & 0.366 \\
Reweight~\cite{reweihgt}          & 0.726 & 0.465 & 0.369 \\
Mixup~\cite{mixup}              & 0.723 & 0.477  & 0.377  \\
CDR~\cite{cdr}                 & 0.733 & 0.454 & 0.363  \\
Garg~\cite{garg}               & 0.729 & 0.498 & 0.247 \\
\midrule
Co-teaching~\cite{co_teaching}                             & 0.536 & 0.746 & 0.362 \\
\rowcolor{gray!15}Co-teaching + Ours    & {\bf 0.737} & {\bf0.434} & {\color{blue}\bf0.388}  \\
\midrule
JoCor~\cite{jocor}                                   & 0.496 & 0.749 & 0.361 \\
\rowcolor{gray!15} JoCor + Ours         & {\bf0.642} & {\bf0.546} & {\color{red}\bf0.393} \\
\midrule
CoDis~\cite{codis}                                   & 0.690 & 0.518 & 0.367 \\
\rowcolor{gray!15}CoDis + Ours          & {\color{red}\bf0.747} & {\color{red}\bf 0.421} & {\bf0.384} \\
\bottomrule
\label{tab:results_dr}
\end{tabular}
\vspace{-5pt}
\end{table}

\section{Conclusion}\label{sec:conclusion}
We proposed a self-relaxed joint-training framework to train a classifier with ordinal noisy labels. The proposed framework is based on the traditional joint-training framework with clean sample selection and dual-network architecture~\cite{co_teaching,jocor,codis} and improved for learning with ordinal noisy labels. Specifically, our framework trains dual networks by soft labels, representing the distribution of true labels. In contrast, ours adheres to using original hard labels for sample selection. This combination of hard and soft labels was essential to improve sample selection accuracy and final classification accuracy. Using three medical image datasets, we confirmed that applying our framework to several state-of-the-art methods within the joint-training framework improved their performance and outperformed other methods for learning with noisy labels.\par

Future work will focus on the {\em class imbalance} problems on ordinal noisy labels. Medical image datasets tend to be class-imbalanced. Unfortunately, the methods for learning with noisy labels suffer from class imbalance ~\cite{imblance_noisy_1,imblance_noisy_2,imblance_noisy_3}. Therefore, it will be helpful for our methods to integrate some techniques to mitigate class imbalance.\par

{\small
\bibliographystyle{ieee_fullname}
\bibliography{egbib}
}

\newpage

\begin{algorithm}[t] 
1: {\bf Input:} Dataset $\tilde{\mathcal{D}}$, two networks $f_1$ and $f_2$ with initialized weights $\bm{\theta}_1$ and $\bm{\theta}_2$, learning rate $\eta$, noise rate $\epsilon$, epoch $T'$ and $T_{\max}$, iteration $t_{\max}$, temperature $\tau$;\\
\For{$T = 1,2,\dots,T_{\max}$}{
	
	2: {\bfseries Shuffle} training set $\tilde{\mathcal{D}}$;\\
	\For{$t = 1,\dots,t_{\max}$}
	{	
		3: {\bfseries Fetch} mini-batch $\tilde{\mathcal{B}}$ from $\tilde{\mathcal{D}}$;\\
            4: {\bfseries Select} clean samples from $\tilde{\mathcal{B}}$ by $\mathcal{L}_\mathrm{h}^\mathrm{JoCor}$ (with $\tau$);\\
            {\hphantom{hoge}$\mathcal{B} \leftarrow \arg\min_{\mathcal{B}':|\mathcal{B}'|\ge R(T)|\tilde{\mathcal{B}}|}\mathcal{L}_\mathrm{h}^\mathrm{JoCor}( f_1, f_2,  \mathcal{B}')$; \hfill} 
            5: {\bfseries Derive} soft labels $\bm{l}_{\mathrm s}$ from $\bm{l}_{\mathrm h}$ for $\mathcal{B}_1, \mathcal{B}_2$ by Eq.({\color{red}3});\\
		6: {\bfseries Update} networks; \\
            \hphantom{hoge}$\bm{\theta}_1 \leftarrow \bm{\theta}_1 - \eta\nabla \mathcal{L}_\mathrm{s}^\mathrm{JoCor}(f_1,f_2,\mathcal{B})$; \hfill \\
		\hphantom{hoge}$\bm{\theta}_2 \leftarrow \bm{\theta}_2 - \eta\nabla \mathcal{L}_\mathrm{s}^\mathrm{JoCor}(f_1,f_2,\mathcal{B})$; \hfill 
	}
	7: {\bfseries Update} $R(T) \leftarrow 1 - \min\left\lbrace  \frac{T}{T'} \epsilon, \epsilon \right\rbrace $;
}
8: {\bf Output:} two trained networks with $\bm{\theta}_1$ and $\bm{\theta}_2$.
\caption{Proposed framework with JoCor~\cite{jocor}}
\label{alg:srjt_jocor}
\end{algorithm}

\begin{algorithm}[t] 
1: {\bf Input:} Dataset $\tilde{\mathcal{D}}$, two networks $f_1$ and $f_2$ with initialized weights $\bm{\theta}_1$ and $\bm{\theta}_2$, learning rate $\eta$, noise rate $\epsilon$, epoch $T'$ and $T_{\max}$, iteration $t_{\max}$, temperature $\tau$;\\
\For{$T = 1,2,\dots,T_{\max}$}{
	
	2: {\bfseries Shuffle} training set $\tilde{\mathcal{D}}$;\\
	\For{$t = 1,\dots,t_{\max}$}
	{	
		3: {\bfseries Fetch} mini-batch $\tilde{\mathcal{B}}$ from $\tilde{\mathcal{D}}$;\\
            4: {\bfseries Select} clean samples from $\tilde{\mathcal{B}}$ by $\mathcal{L}_\mathrm{h}^\mathrm{CoDis}$ (with $\tau$);\\
            {\hphantom{hoge}$\mathcal{B}_1 \leftarrow \arg\min_{\mathcal{B}':|\mathcal{B}'|\ge R(T)|\tilde{\mathcal{B}}|}\mathcal{L}_\mathrm{h}^\mathrm{CoDis}(  f_1, f_2, \mathcal{B}')$; \hfill} \\
            {\hphantom{hoge}$\mathcal{B}_2 \leftarrow \arg\min_{\mathcal{B}':|\mathcal{B}'|\ge R(T)|\tilde{\mathcal{B}}|}\mathcal{L}_\mathrm{h}^\mathrm{CoDis}(  f_2, f_1, \mathcal{B}')$; \hfill} \\
            5: {\bfseries Derive} soft labels $\bm{l}_{\mathrm s}$ from $\bm{l}_{\mathrm h}$ for $\mathcal{B}_1, \mathcal{B}_2$ by Eq.({\color{red}3});\\
		6: {\bfseries Update} networks; \\
            \hphantom{hoge}$\bm{\theta}_1 \leftarrow \bm{\theta}_1 - \eta\nabla \mathcal{L}_\mathrm{s}(f_1,\mathcal{B}_2)$; \hfill \\
		\hphantom{hoge}$\bm{\theta}_2 \leftarrow \bm{\theta}_2 - \eta\nabla \mathcal{L}_\mathrm{s}(f_2,\mathcal{B}_1)$; \hfill 
	}
	7: {\bfseries Update} $R(T) \leftarrow 1 - \min\left\lbrace  \frac{T}{T'} \epsilon, \epsilon \right\rbrace $;
}
8: {\bf Output:} two trained networks with $\bm{\theta}_1$ and $\bm{\theta}_2$.
\caption{Proposed framework with CoDis~~\cite{codis}}
\label{alg:srjt_codis}
\end{algorithm}

\begin{table*}[t]
\small
\centering
\caption{Classification results on \underline{LIMUC} with \underline{Truncated-Gaussian noise}. Following tradition, the test accuracy (Acc.), mean absolute error (MAE), and macro F1 (mF1) are averaged over the last ten epochs. The mean and standard deviations of five-fold cross-validation are shown. The best and second-best results are highlighted in \textcolor{red}{red}\ and  \textcolor{blue}{{blue}}, respectively. For plugin settings, improved results are shown by {\bf bold}.}
\begin{tabular}{lcccccc}
\toprule
\multirow{2}{*}{Method} & \multicolumn{3}{c}{Noise rate: $\epsilon=0.2$} & \multicolumn{3}{c}{Noise rate: $\epsilon=0.4$} \\
\cmidrule(lr){2-4} \cmidrule(lr){5-7} & Acc.$\uparrow$  & MAE$\downarrow$ & mF1$\uparrow$ & Acc.$\uparrow$ & MAE$\downarrow$ & mF1$\uparrow$ \\
\midrule
Standard            & 0.665$\pm$0.010 & 0.373$\pm$0.007 & 0.573$\pm$0.007 & 0.566$\pm$0.018 & 0.489$\pm$0.018 & 0.479$\pm$0.011 \\
Sord~[{\color{green}{5}}]                & 0.708$\pm$0.009 & 0.309$\pm$0.010 & 0.632$\pm$0.015 & 0.632$\pm$0.016 & 0.389$\pm$0.019 & 0.564$\pm$0.024 \\
Label-smooth~[{\color{green}{23}}]  & 0.690$\pm$0.010 & 0.339$\pm$0.010 & 0.601$\pm$0.016 & 0.609$\pm$0.016 & 0.432$\pm$0.017 & 0.511$\pm$0.007 \\
F-correction~[{\color{green}{25}}]       & 0.670$\pm$0.009 & 0.362$\pm$0.010 & 0.585$\pm$0.010 & 0.609$\pm$0.008 & 0.430$\pm$0.008 & 0.529$\pm$0.010 \\
Reweight~[{\color{green}{18}}]           & 0.667$\pm$0.006 & 0.371$\pm$0.008 & 0.573$\pm$0.013 & 0.575$\pm$0.008 & 0.477$\pm$0.008 & 0.494$\pm$0.013 \\
Mixup~[{\color{green}{9}}]              & 0.676$\pm$0.008 & 0.359$\pm$0.005 & 0.583$\pm$0.011 & 0.605$\pm$0.012 & 0.449$\pm$0.015 & 0.490$\pm$0.013 \\
CDR~[{\color{green}{39}}]                 & 0.674$\pm$0.012 & 0.362$\pm$0.007 & 0.582$\pm$0.016 & 0.571$\pm$0.027 & 0.482$\pm$0.027 & 0.481$\pm$0.015 \\
Garg~[{\color{green}{7}}]               & 0.657$\pm$0.054 & 0.433$\pm$0.146 & 0.447$\pm$0.128 & 0.525$\pm$0.040 & 0.786$\pm$0.121 & 0.267$\pm$0.015 \\
\midrule
Co-teaching~[{\color{green}{8}}]                             & 0.698$\pm$0.002 & 0.332$\pm$0.004 & 0.610$\pm$0.012 & 0.646$\pm$0.020 & 0.393$\pm$0.023 & 0.544$\pm$0.023 \\
\rowcolor{gray!15}Co-teaching + Ours    & {\color{red}\bf0.731$\pm$0.005} & {\color{blue}\bf0.289$\pm$0.005} & {\color{red}\bf0.646$\pm$0.014} & {\bf 0.677$\pm$0.019} & {\bf 0.356$\pm$0.019} & {\bf0.545$\pm$0.011} \\
\midrule
JoCor~[{\color{green}{35}}]                                   & 0.720$\pm$0.006 & 0.306$\pm$0.006 & 0.633$\pm$0.008 & \color{red}{0.690$\pm$0.015} & \color{blue}{0.345$\pm$0.017} & \color{blue}{0.573$\pm$0.007} \\
\rowcolor{gray!15} JoCor + Ours         & {\color{blue}\bf0.731$\pm$0.009} & {\color{red}\bf0.287$\pm$0.010} & {\bf\color{blue}{0.642$\pm$0.018}} & 0.678$\pm$0.016 & 0.353$\pm$0.017 & 0.549$\pm$0.009 \\
\midrule
CoDis~[{\color{green}{37}}]                                   & 0.694$\pm$0.004 & 0.336$\pm$0.005 & 0.609$\pm$0.013 & 0.622$\pm$0.012 & 0.418$\pm$0.013 & 0.530$\pm$0.014 \\
\rowcolor{gray!15}CoDis + Ours          & {\bf 0.723$\pm$0.005} & {\bf 0.294$\pm$0.006} & {\bf 0.639$\pm$0.017} & {\color{blue}\bf 0.684$\pm$0.012} & {\color{red}\bf 0.342$\pm$0.015} & {\color{red}\bf 0.581$\pm$0.028} \\
\bottomrule
\label{tab:results_limuc_neighbor}
\end{tabular}
\end{table*}

\begin{table*}[t]
\small
\centering
\caption{Results of \underline{LIMUC dataset} with \underline{Truncated-Gaussian noise} under different loss usages for sample selection and updating. The best and second-best results are highlighted in \textcolor{red}{red}\ and  \textcolor{blue}{{blue}}, respectively. \label{tab:abl-limuc-neighbor}}
\begin{tabular}{cccccccc}
\toprule
\multirow{2}{*}{Selection} & \multirow{2}{*}{Updating} &\multicolumn{3}{c}{Noise rate: $\epsilon=0.2$} & \multicolumn{3}{c}{Noise rate: $\epsilon=0.4$} \\
\cmidrule(lr){3-5} \cmidrule(lr){6-8} & & Acc.$\uparrow$  & MAE$\downarrow$ & mF1$\uparrow$ & Acc.$\uparrow$ & MAE$\downarrow$ & mF1$\uparrow$ \\
\midrule
hard & hard & 0.698$\pm$0.002 & 0.332$\pm$0.004 & 0.610$\pm$0.012 & 0.646$\pm$0.020 & 0.393$\pm$0.023 & {\color{blue}0.544$\pm$0.023} \\
soft & soft & {\color{blue}0.722$\pm$0.006} & {\color{blue}0.300$\pm$0.008} & {\color{blue}0.628$\pm$0.019} & {\color{blue}0.661$\pm$0.021} & {\color{blue}0.382$\pm$0.024} & 0.489$\pm$0.021 \\
\rowcolor{gray!15}hard & soft & {\color{red}0.731$\pm$0.005} & {\color{red}0.289$\pm$0.005} & {\color{red}0.646$\pm$0.014} & {\color{red}0.677$\pm$0.019} & {\color{red}0.356$\pm$0.019} & {\color{red}0.545$\pm$0.011} \\
\bottomrule
\end{tabular}
\bigskip\\
\small
\centering
\caption{Results of \underline{our private dataset} with \underline{Truncated-Gaussian noise} under different loss usages for sample selection and updating.\label{tab:abl-private-neighbor}}
\begin{tabular}{cccccccc}
\toprule
\multirow{2}{*}{Selection} & \multirow{2}{*}{Updating} &\multicolumn{3}{c}{Noise rate: $\epsilon=0.2$} & \multicolumn{3}{c}{Noise rate: $\epsilon=0.4$} \\
\cmidrule(lr){3-5} \cmidrule(lr){6-8} & & Acc.$\uparrow$  & MAE$\downarrow$ & mF1$\uparrow$ & Acc.$\uparrow$ & MAE$\downarrow$ & mF1$\uparrow$ \\
\midrule
hard & hard & 0.788$\pm$0.009 & 0.236$\pm$0.008 & 0.599$\pm$0.031 & 0.702$\pm$0.022 & 0.328$\pm$0.020 & \color{blue}{0.490$\pm$0.036} \\
soft & soft & {\color{blue}0.809$\pm$0.007} & {\color{blue}0.209$\pm$0.006} & {\color{blue}0.611$\pm$0.028} & {\color{blue}0.721$\pm$0.030} & {\color{blue}0.318$\pm$0.030} & 0.442$\pm$0.038 \\
\rowcolor{gray!15}hard & soft      & {\color{red}0.815$\pm$0.010} & {\color{red}0.202$\pm$0.008} & {\color{red}0.621$\pm$0.035} & {\color{red}0.748$\pm$0.031} & {\color{red}0.282$\pm$0.033} & {\color{red}0.491$\pm$0.032} \\
\bottomrule
\end{tabular}
\end{table*}

\subsection*{Appendix A.\ Applying our framework to other joint-training methods}
In our paper, we detailed Algorithm~\ref{alg:srjt}, where our framework is applied to Co-teaching~\cite{co_teaching}. However, our framework is versatile and can be applicable to other joint-training methods, such as JoCor~\cite{jocor} and CoDis~\cite{codis}. Algorithms \ref{alg:srjt_jocor} and \ref{alg:srjt_codis} show the entire training procedure of ``JoCor + Ours'' and ``CoDis + Ours,'' respectively. 
\par

Algorithm~\ref{alg:srjt_jocor} of ``JoCor + Ours'' has a very similar structure as Algorithm~\ref{alg:srjt}; however, its loss functions $\mathcal{L}_\mathrm{h}^\mathrm{JoCor}$
and  $\mathcal{L}_\mathrm{s}^\mathrm{JoCor}$ are different from $\mathcal{L}_\mathrm{h}$ and $\mathcal{L}_\mathrm{s}$, respectively. JoCor~\cite{jocor} uses the common clean sample set $\mathcal{B}$ for the two networks and introduces co-regularization to reduce divergence between the networks. Consequently, $\mathcal{L}_\mathrm{h}^\mathrm{JoCor}$ becomes:
\begin{eqnarray}
\lefteqn{\mathcal{L}_\mathrm{h}^\mathrm{JoCor}(f_1,f_2,\tilde{\mathcal{B}})=\nonumber} \\ 
&&  (\mathcal{L}_\mathrm{h}(f_1,\tilde{\mathcal{B}}) + \mathcal{L}_\mathrm{h}(f_2,\tilde{\mathcal{B}})) + \lambda\mathcal{L}_\mathrm{reg}(f_1,f_2,\tilde{\mathcal{B}}),
\label{eq:L_h_jocor}
\end{eqnarray}
where  $\mathcal{L}_\mathrm{reg}$ is a regularization term:
\begin{align}
\mathcal{L}_\mathrm{reg}(f_1,f_2,\tilde{\mathcal{B}}) =  \sum_{\{\bm{x}_i,\tilde{y}_i\} \in \tilde{\mathcal{B}}} J(\bm{p}_1(\bm{x}_i), \bm{p}_2(\bm{x}_i)), 
\end{align}
and $J(\cdot,\cdot)$ denotes the Jeffrey divergence (i.e., the symmetrized Kullback-Leibler (KL) divergence). For updating the models with soft labels, ``JoCor+Ours'' uses the loss function $\mathcal{L}_\mathrm{s}^\mathrm{JoCor}$ obtained by replacing $\mathcal{L}_\mathrm{h}$ with $\mathcal{L}_\mathrm{s}$ in Eq.~(\ref{eq:L_h_jocor}).
\par

Algorithm~\ref{alg:srjt_codis} of ``CoDis + Ours'' has a more elaborated structure than Algorithm~\ref{alg:srjt};  CoDis~\cite{codis} uses possibly clean samples that have high discrepancy prediction probabilities between two networks, $f_1$ and $f_2$. The proposed framework with CoDis selects small loss samples with the loss function:
\begin{align}
\mathcal{L}_\mathrm{h}^\mathrm{CoDis}(f_1,f_2,\tilde{\mathcal{B}}) = \mathcal{L}_\mathrm{h}(f_1,\tilde{\mathcal{B}}) - \lambda\mathcal{L}_\mathrm{reg}(f_1,f_2,\tilde{\mathcal{B}}).
\label{eq:L_h_codis}
\end{align}
For updating the models with soft labels, ``CoDis + Ours'' uses the loss function $\mathcal{L}_\mathrm{s}$.

\begin{figure}[t]
    \centering
    \includegraphics[width=\linewidth]{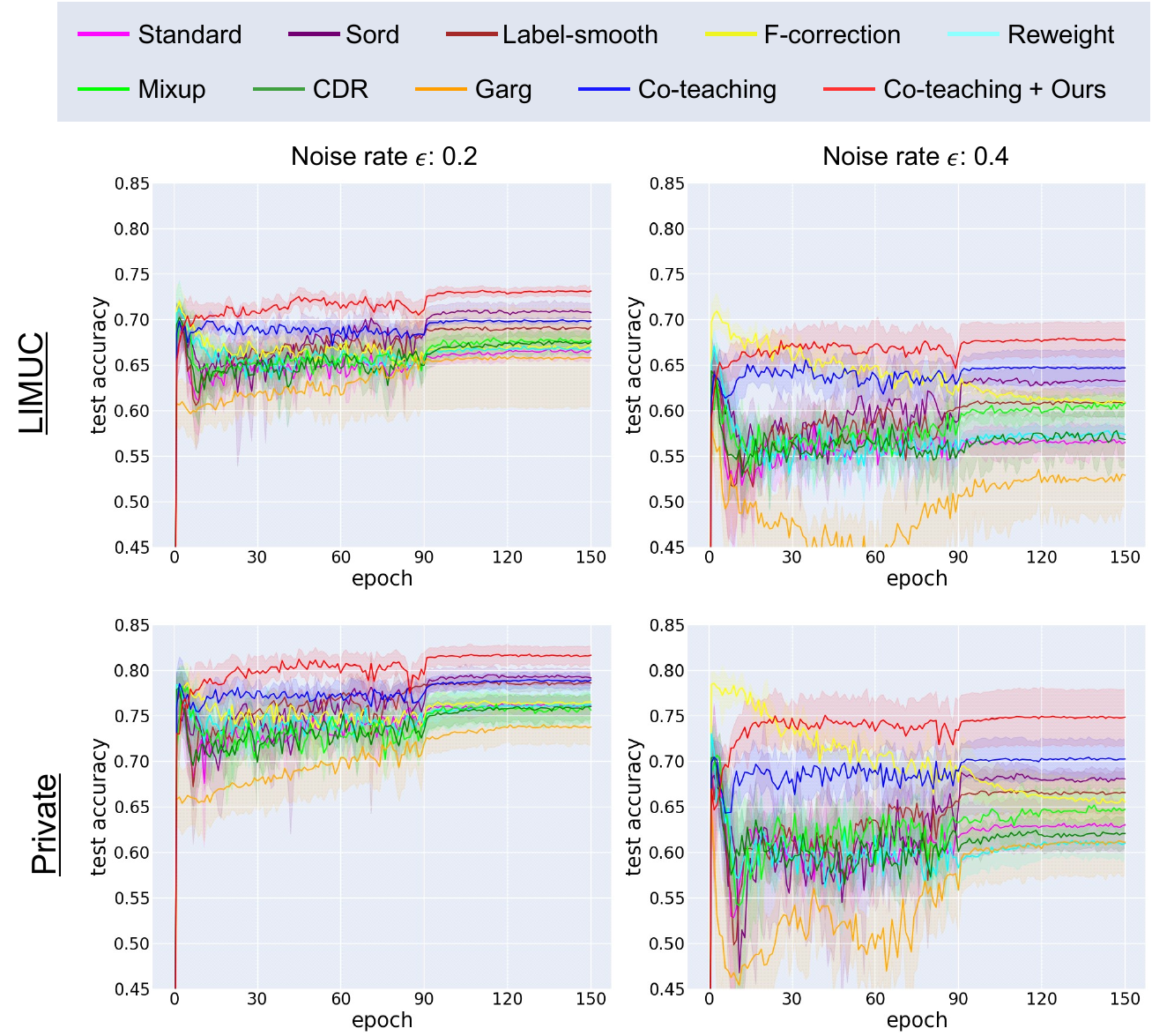}
    \caption{Test accuracy curves. 
    The width of the shading indicates the standard deviation in cross-validation.}
    \label{fig:test_acc_supp}
\end{figure}

\begin{figure}[t]
    \centering
    \includegraphics[width=\linewidth]{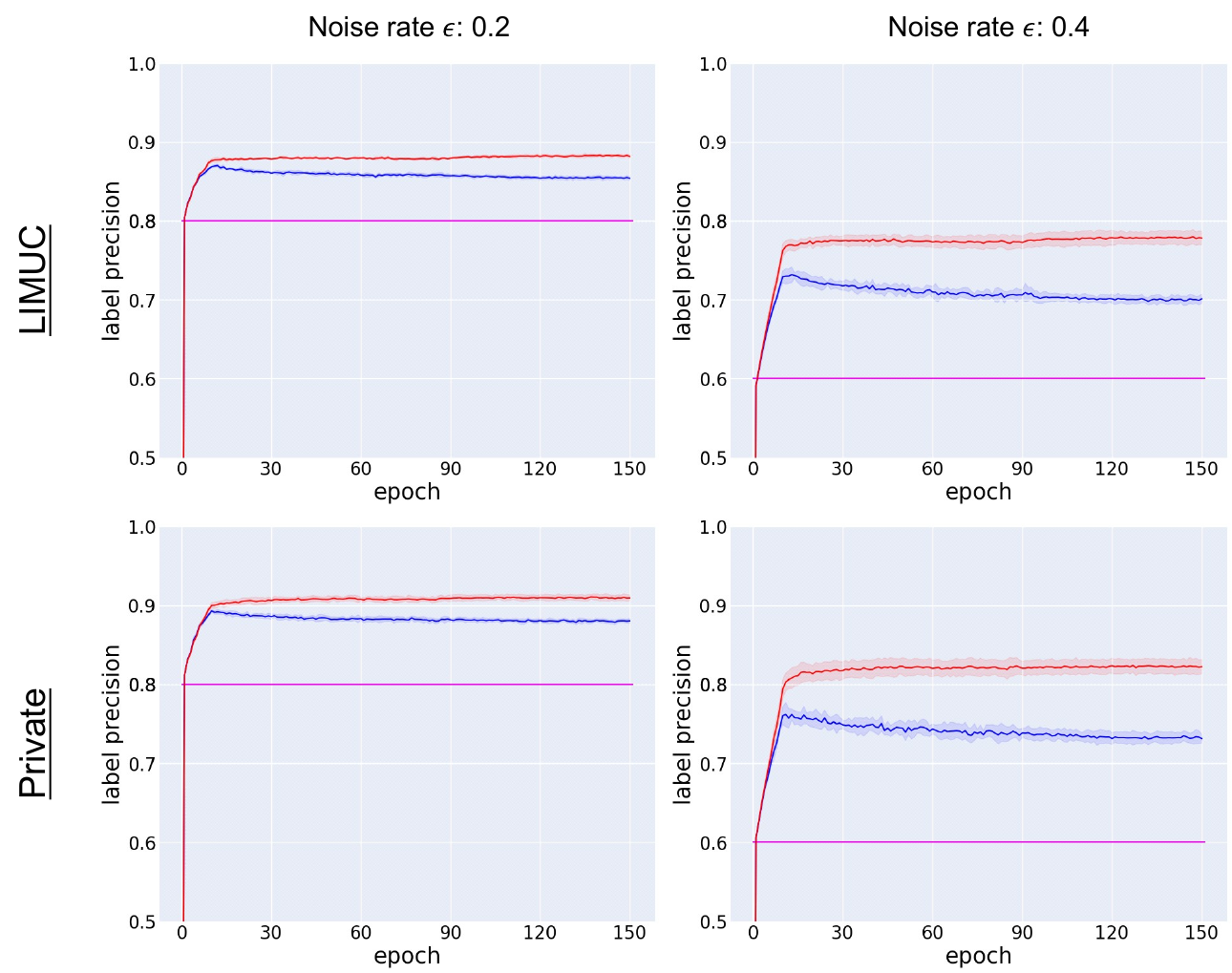}
    \caption{Label precision curves. The \blue{blue} and \red{red} curves show the label precisions by ``Co-teaching'' and ``Co-teaching + Ours,'' respectively. The pink horizontal line shows $(1-\epsilon)$.}
    \label{fig:label_precision_supp}
\end{figure}

\subsection*{Appendix B.\ Experimental evaluations under the Truncated-Gaussian noise}
Table~\ref{tab:results_limuc_neighbor} shows the results on LIMUC~\cite{limuc} under the Truncated-Gaussian noise, simulating the case that experts make the mis-labelings between the neighboring labels. (Specifically, the $i,j$th element of the label transition matrix,$P_{ij}$, takes $1 - \rho$ for $|i-j|=1$ and $P_{ij} = 0$ for $|i-j|>1$.) Our methods (``$*$ + Ours'') outperform the others. Compared to the results under the Quasi-Gaussian noise, the individual accuracies in Table~\ref{tab:results_limuc_neighbor} are slightly lower, which is the same trend seen in the results on our private dataset in Section~\ref{ssec:results}. 

Tables~\ref{tab:abl-limuc-neighbor} and \ref{tab:abl-private-neighbor} show how the combination of $\mathcal{L}_\mathrm{h}$ and $\mathcal{L}_\mathrm{s}$ is appropriate for learning with ordinal noisy labels under Truncated-Gaussian. These tables show the results for LIMUC and the private dataset, respectively. The tendency of the results is almost the same as those under the Quasi-Gaussian noise, shown in Tables~\ref{tab:abl-limuc} and \ref{tab:abl-private}.

Fig.~\ref{fig:test_acc_supp} shows the test accuracy curves for the individual methods on two UC datasets with Truncated-Gaussian noise. The comparative methods show a sharp increase in their test accuracy in early epochs. Then, the comparative models often start ``memorizing'' the samples with incorrect labels. Our method (``Co-teaching + Ours'') could avoid the memorization effect.

Fig.~\ref{fig:label_precision_supp} shows the change in label precision on two UC datasets with Truncated-Gaussian noise. The backbone method is Co-teaching. The pink horizontal lines ($1-\epsilon$) indicate the label precision under random sample selection. Our method (``Co-teaching + Ours,''  the red curve) shows far better label precisions than random selection (pink line) and Co-teaching (the blue curve).  

\section*{Appendix C.\ Code avalilability}
We share our codes at \url{https://github.com/shumpei-takezaki/Self-Relaxed-Joint-Training}.

\end{document}